\documentclass[sigconf]{acmart}
\pdfoutput=1
\renewcommand\footnotetextcopyrightpermission[1]{}
\setcopyright{none}

\def\BibTeX{{\rm B\kern-.05em{\sc i\kern-.025em b}\kern-.08emT\kern-.1667em\lower.7ex\hbox{E}\kern-.125emX}}

\usepackage[utf8]{inputenc}
\usepackage{todonotes}
\presetkeys{todonotes}{inline}{}
\usepackage{float}
\usepackage[caption = false]{subfig}
\usepackage{mathtools}

\begin{document}

\title[Automating concept-drift detection]{Automating concept-drift detection \\ by self-evaluating predictive model degradation}

\author{Tania Cerquitelli, Stefano Proto, Francesco Ventura, Daniele Apiletti, Elena Baralis}
\affiliation{%
    \institution{Department of Control and Computer Engineering \\Politecnico di Torino}
    \city{Turin}
    \state{Italy}
}
\email{name.surname@polito.it}
\renewcommand{\shortauthors}{Cerquitelli, et al.}

\begin{abstract}
A key aspect of automating predictive machine learning entails the capability of properly triggering the update of the trained model. 
To this aim, suitable automatic solutions to self-assess the prediction quality and the data distribution drift between the original training set and the new data have to be devised.

In this paper, we propose a novel methodology to automatically detect prediction-quality degradation of machine learning models due to class-based concept drift, i.e., when new data contains samples that do not fit the set of class labels known by the currently-trained predictive model.

Experiments on synthetic and real-world public datasets show the effectiveness of the proposed methodology in automatically detecting and describing concept drift caused by changes in the class-label data distributions.

\end{abstract}


\maketitle
\section{Introduction}
\label{sec:introduction}

In many application domains, from industrial production environments to smart cities, from network traffic classification to text mining, it is quite common that the nature of the collected data changes over time due to the evolution of the phenomena under analysis, e.g., equipment maintenance, road topology changes, updated configurations, and environmental factors. 
However, collecting historical training sets including all possible class labels may be very hard, too expensive, or even unfeasible. 
Hence, when performing predictions, new data belonging to unseen class labels, i.e., labels that the predictive model did not know at training time, are likely to be presented at some point in the future, leading to wrong predictions.

Frequently updating the prediction models by extending the training set to the new data can be computationally intensive and, worse, may require the intervention of domain experts along with data scientists to translate the changes of the phenomenon into appropriate choices for the predictive task.
For these reasons, it is often unfeasible or at least highly sub-optimal to simply retrain the model very frequently.

To this aim, we introduce a self-evaluation step to automatically identify and quantify the degradation of the prediction quality over time.
An additional challenge in this approach is provided by the absence of the ground-truth predictions for the newly classified samples. Hence, the solution is based on \textit{unsupervised} evaluation metrics.
These metrics can provide a twofold contribution: (i)~automatically trigger a predictive model retraining, by determining when it is required to better fit the new data, and (ii)~describe the changes in data distributions motivating the model update.

Many works addressed the problem of detecting data distribution changes in machine learning.
In \cite{tsymbal2004problem}, three different strategies are presented: 
(i) instance selection, 
(ii) instance weighting and 
(iii) ensemble learning.
In the first case, the goal is to consider samples relevant to the current concept: these approaches are mainly based on trailing windows moving over the latest instances.
The instance-weighting strategy is based on learning algorithms able to consider weights separately for each instance, whereas the latter exploits ensembles to better capture different nuances in the data.
In our solution, we exploit a trailing window on new data samples similarly to strategy~(i), but we propose a new evaluation metrics to self-evaluate degradation between the training set and the current dataset, since the training set describes what the current predictive model has learnt.

A survey on concept drift adaptation is provided in \cite{gama2014survey}. In this context, we focus on class-based concept drift for predictive analytics, for which a unifying view is provided and its nuances are explored in \cite{moreno2012unifying}.
Different techniques for monitoring two distributions at different time windows have been proposed.
An entropy-based measure for data streams has been proposed in \cite{4053163} to detect abrupt concept drifts due to context changes produced by sensors (e.g., wearable devices).
In \cite{bifet2007learning} the authors propose an adaptive sliding window technique along with a Naïve Bayes predictor to monitor the performance of a predictive model over time. 
Another approach addressing concept drift detection in textual data, based on support vector machines, is presented in \cite{klinkenberg2000detecting}.
An analysis of discrete-time Markov chains affected by concept drift has been recently proposed in \cite{8618443} providing a collection of change-detection mechanisms and an adaptive learning algorithm suited to this specific context.
Recent approaches aiming at detecting concept drift in the context of online learning with imbalanced classes are presented in \cite{8246564}, whereas the challenge of detecting model degradation when incremental learning is applied was addressed in \cite{8246541}.
Many related works validate their approaches on controlled synthetic datasets with known concept drifts, besides considering real-world datasets.

The solution presented in this paper improves the state-of-the-art by defining a methodology for detecting concept drifts on new incoming data over time
(i) automatically (self-evaluation), 
(ii) in a general purpose manner (not tailored to a specific use case or application domain, nor to a specific data type), and
(iii) with scalability in mind. The algorithm has been designed to be horizontally scalable on Big Data contexts by means of the Map Reduce programming paradigm and is implemented on top of Apache Spark.

The paper is organized as follows. 
Section \ref{sec:solution} introduces the overall methodology to automatically provide up-to-date predictive models when required. 
Section \ref{sec:approach} presents the new self-evaluation strategy proposed to detect predictive degradation and concept drift, while Section \ref{sec:experimental-results} discusses some preliminary experiments to evaluate the proposed approach. 
Section \ref{sec:conclusion} draws conclusions and discusses future extensions of the research work.
\section{Automated concept drift management}

\label{sec:solution}
We present a novel unsupervised methodology able to automatically detect class-based concept drifts by evaluating the degradation of the predictions.
Specifically, we aim at identifying when additional or different class labels are required because the current ones misrepresent the new data samples. 

\begin{figure}[!h!]
    \centering
    \includegraphics[width=0.38\textwidth]{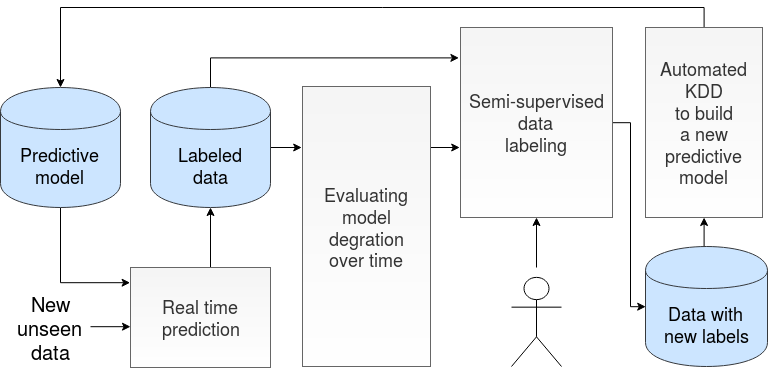}
    \caption{Building blocks of the proposed framework.}
    \label{fig:framework} 
\end{figure}

The proposed methodology consists of three steps, as depicted in Figure \ref{fig:framework}.
(i)~\textit{Model degradation self-evaluation} step, performed through a novel unsupervised approach assessing the degradation of a prediction model over time.
(ii)~\textit{Semi-supervised data labeling} step, 
to assign labels to the new automatically-discovered classes of data. 
A small subset of representative samples of the new classes will be manually inspected by domain experts. Their label assignments will be used for the remaining samples in each corresponding class.
(iii)~\textit{Automated KDD (Knowledge Discovery Process) to build a new predictive model}, able to correctly fit the new incoming data distributions and classification labels. 
This step can be automatically triggered based on the results of the previous ones, e.g., when the model degradation is higher than a given threshold.
The latter step has already seen applications of state-of-the-art approaches tailored to specific data types, such as
in \cite{DBLP:conf/ispa/ApilettiBCMMPV18} for predictive maintenance, 
in \cite{DBLP:conf/sac/CorsoCV17, DBLP:conf/bigdata/ProtoCVC18} for addressing topic detection among textual document collections, 
and for network traffic characterization in \cite{apiletti2016selina}.

The contribution of this paper focuses on step (i), which is a core step required to provide a solution for self-evaluating the model degradation over time, aimed at detecting concept drifts.
The contribution is detailed in Section \ref{sec:approach}.
\section{Self-evaluating model degradation}
\label{sec:approach}

The knowledge of a prediction model is based on the information learnt from the training samples (historical data with labels). Hence, predictions for new data describing evolving phenomena may become misleading or erroneous, due to the different data distributions of the new samples.
To efficiently capture this model degradation over time,
the proposed approach performs  
(i)~\textit{Baseline computation,} by computing unsupervised quality metrics on the training set, and 
(ii)~\textit{Self-evaluation,} by periodically recomputing the same metrics on the new data and comparing it to the metrics in (i).
While new data are continuously classified by the model over time, the quality assessment has to be performed periodically. 
Since the quality metrics depend on the cardinality of the new data and their labels, in the current version we trigger self-evaluation when any class label is affected by a percentage increase in the number of its new samples with respect to the previous run above a given threshold (e.g., 20\%).

Finally, to fully automate the process, a degradation threshold should be defined to trigger predictive model rebuilding. This threshold depends on the expectations of domain experts and end-users, the risks and costs related to the specific application, and also the number of records and classes of the dataset. Its evaluation is out of the scope of this paper.

\textbf{Self-evaluating predictive model quality over time.}
Traditional evaluation techniques for predictive analytics (e.g., f-measure, precision, recall) are not applicable to our context since they require ground-truth labeled data, which is missing for newly-classified samples.
Hence, we exploit an unsupervised index that, given a dataset of samples divided into classes (labels), is able to quantify both the intra-class cohesion and the inter-class separation.
The degradation is then defined as a negative change in the index value of the newly-classified data with respect to the value obtained on the training set, considered as a baseline.

The proposed approach is independent of a specific quality index selection. In this work we exploit the Silhouette \cite{Rousseeuw1987}, a succinct measure of the fit of each sample within its predicted class. It measures how similar a sample is to its own class (cohesion) compared to other classes (separation). 
The Silhouette value ranges from -1 to +1, with higher values indicating a better match to the assigned class and a poor match to other classes.
In order to compute the Silhouette, for each sample, the pairwise distance between itself and each other sample of the dataset have to be calculated. 
For this reason, the traditional computation of the Silhouette coefficient is not suitable for Big Data scenarios. 
However, scalable approaches have been proposed in literature. We adopt the approach proposed in \cite{ventura2019scale}. 
The Silhouette measure can be used for any data type by using the appropriate distance-similarity measure (e.g., Euclidean distance for structured and numerical data, cosine similarity for textual data, Jaccard for Boolean data \cite{Tan:2005:IDM:1095618}).

\textbf{Estimating model degradation.}
To this aim, we compare two quality index values: 
(i) the baseline, computed on the training set only, and
(ii) the current one, computed on newly-labeled data.
The change in quality has to be quantified separately for each class.
Hence, at each index computation step, for each class, a curve is plotted, representing the Silhouette values of each point, sorted in increasing order.
To enable pairwise Silhouette curve comparison, the two curves should be characterized by the same number of points. Hence, a down-sampling of the curve characterized by the largest cardinality may be required.

An upward shift of the Silhouette curve represents an improvement in terms of intra-class cohesion and inter-class separation, while, on other hand, a downward shift denotes a degradation.
The degradation of the Silhouette curve is able to detect the presence of new samples not fitting the distribution of the data seen at model-training time.
Hence, it is likely that the current prediction model is not able to correctly assign the class labels to these new samples. Furthermore, the correct class labels might be new (additional) ones, unknown for the current model.

To quantify the shift of the Silhouette curves, the MAAPE (\textit{Mean Arc-tangent Absolute Percentage Error}) \cite{MAAPE-kim2016new} has been used, although other error metrics could be used alternatively. 
Given a prediction model trained on a set of classes $C$ at time $t_0$, whose training set has a Silhouette value $Sil_{t_0}$, the degradation of a class $c \in C$ at time $t$ is described by the following relation:

\begin{gather}
    DEG(c, t) = \alpha * \textrm{MAAPE}(Sil_{t_0}, Sil_t) * \frac{N_c}{N} 
    \label{eq:degradation}\\
    \alpha = \left\{\begin{matrix}
    1 & if:\ \overline{Sil_{t_0}} \geq \overline{Sil_t}
    \\ 
    -1 & if:\ \overline{Sil_{t_0}} < \overline{Sil_t}
    \end{matrix}\right.
    \label{eq:alpha-def}
\end{gather}
The coefficient $\alpha$ defines if the degradation is positive, meaning a possible reduction in performance of the prediction model, or negative, when the new data fit the training distribution, hence increasing the cohesion of class $c$.
From (\ref{eq:degradation}), the degradation is modelled as the MAAPE error between the baseline Silhouette $Sil_{t_0}$ on the training set and the possibly-degraded Silhouette $Sil_{t}$ computed at time $t$ on newly-labeled data.
The degradation estimation is weighted by the ratio ${N_c}/{N}$, where $N_c$ is the number of new records assigned to class $c$ and $N$ is the total number of new samples, with $N$ capped at the number of samples in the training set $N_{train}$.
The cap is introduced to allow a fair comparison when $N>>N_{train}$
To this aim, $Sil_{t}$ is computed on a trailing window containing the latest $N$ samples up to the number of samples of the training set $N_{train}$.
Finally, the degradation of the whole model at time $t$ is computed as the sum of the degradation values for each class: 
$\sum_{c \in C} DEG(c, t)$.

As a heuristics to automate the degradation assessment, we are currently triggering the self-evaluation when at least one class has seen an increase of a given percentage in $N_c$ with respect to the latest computation (e.g., 20\%). 
This heuristics allows us to avoid delaying the detection of concept drift by providing excessive inertia to the approach.
Finally, we are considering to trigger a full model rebuild when the overall degradation or at least a single-class degradation are above given thresholds (e.g., 10-15\% overall and 5-10\% single class). The assessment of the thresholds and the evaluation of different heuristics will be addressed as future work.

\section{Preliminary experimental results}
\label{sec:experimental-results}

We present experimental results on two datasets containing concept shifts represented by the presence of data belonging to previously unseen classes, unknown to the predictive model. 
The first datataset, D1, is a synthetic dataset created with the \textit{scikit-learn} Python library \cite{scikit-learn}. It has been generated with 4 normally distributed classes and 800,000 records, 200,000 for each class, and 10 features.
Dataset D2, on the contrary, is a real-world dataset containing Wikipedia articles, extracted from 3 classes: 
\textit{mathematics}, 
\textit{literature}, and \textit{food-drink}. 
For each class, a selection of 1,000 articles appearing in the Wikipedia index for that class have been downloaded. 
Each article has been pre-processed to obtain an embedded representation of 100 features for it.
The document-embedding process takes advantage of a Doc2Vec model \cite{le2014distributed}, pre-trained on the English Wikipedia corpus \cite{lau2016empirical}.
For all datasets, a Random Forest classifier has been used as predictive model. 
Using a 3-fold cross-validation, the average f-measure of the predictive model is 0.964 for dataset D1, and 0.934 for dataset D2.

\begin{figure}
    \centering
    \subfloat[Degradation with class 2.]{
    \includegraphics[width=0.34\textwidth]{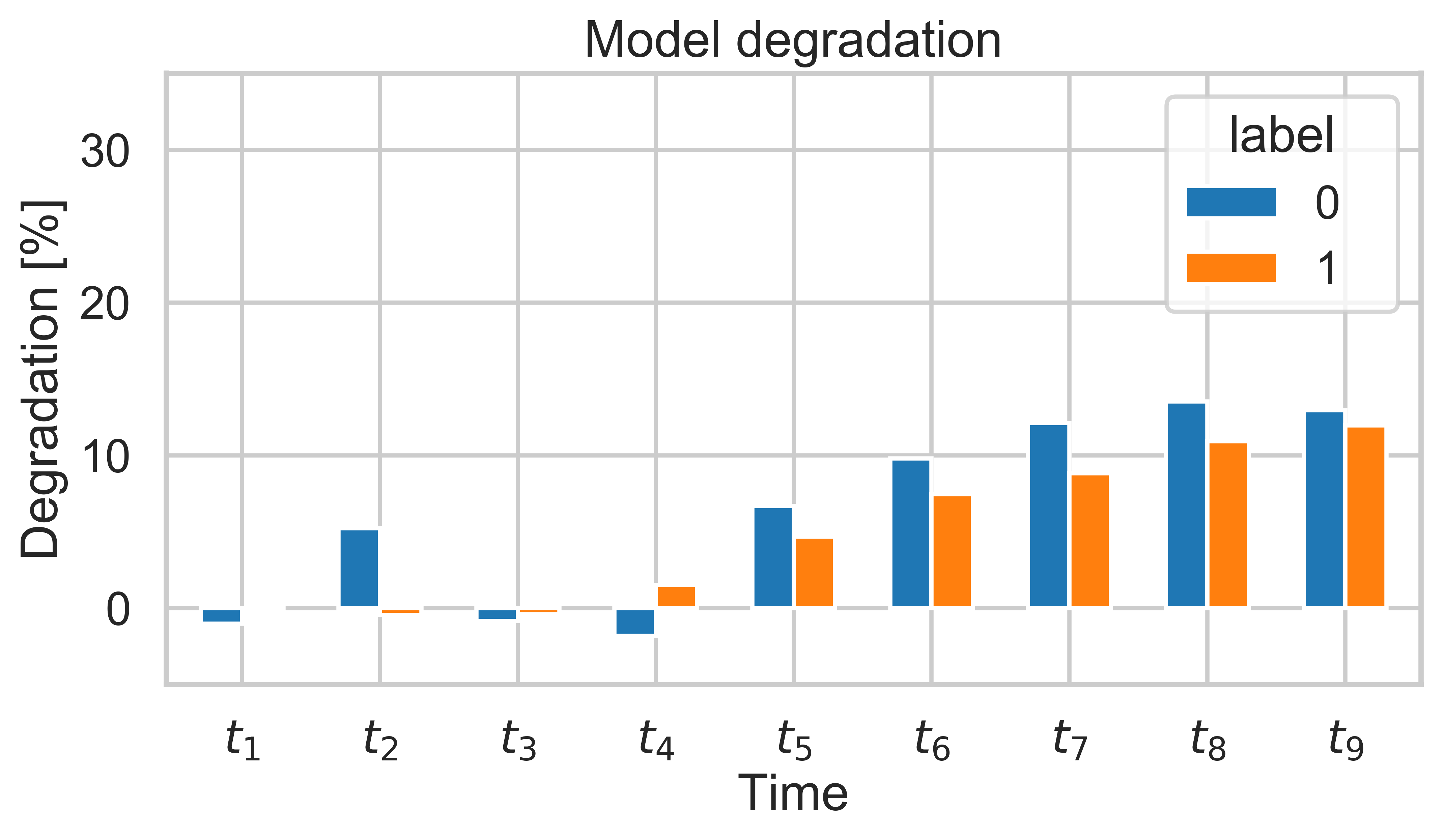}
    \label{fig:model-degradation-over-time-1} 
    }\\
    \subfloat[Degradation with class 3.]{
    \includegraphics[width=0.34\textwidth]{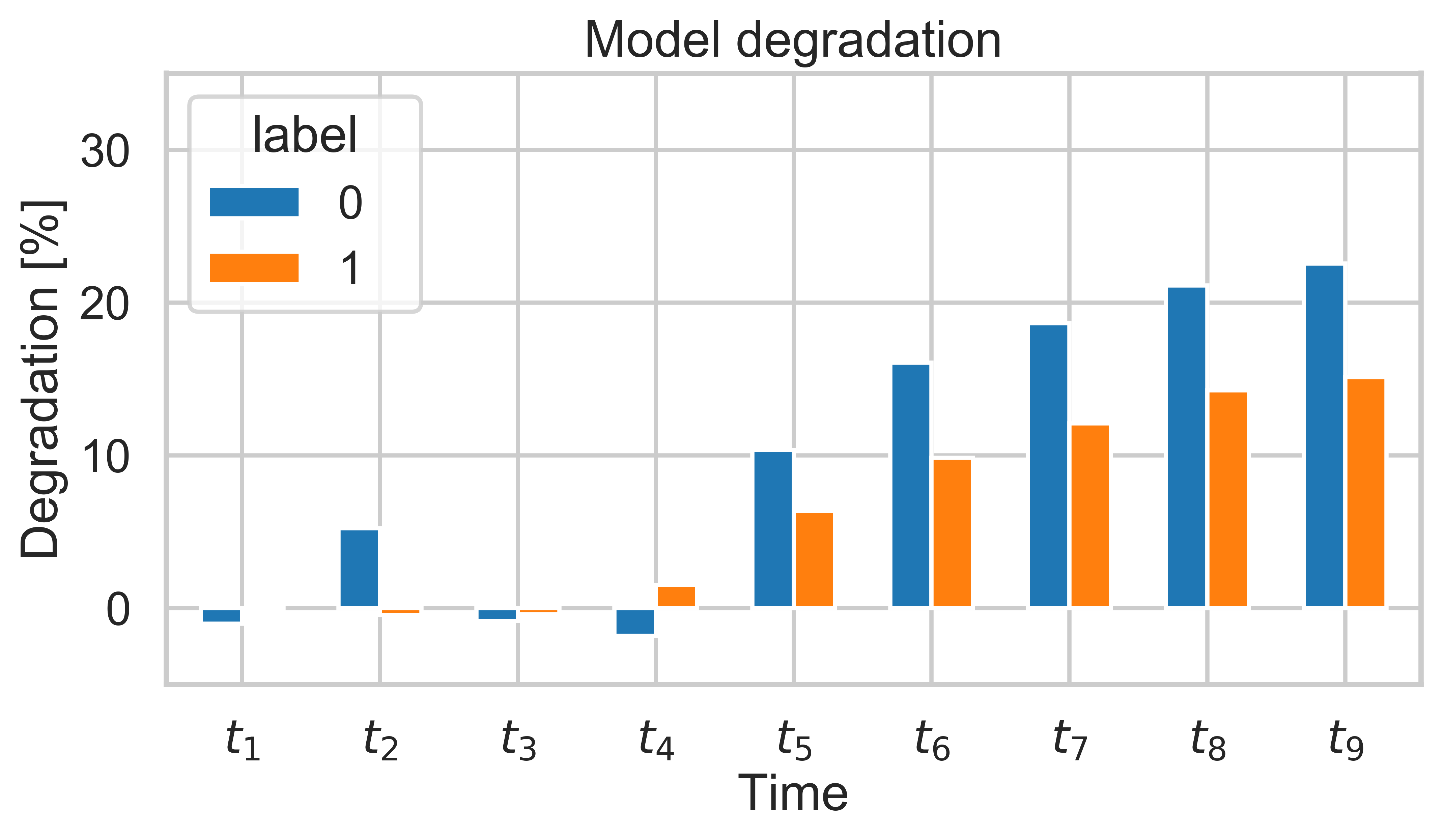}
    \label{fig:model-degradation-over-time-2}
    }
    \caption{Dataset D1. Model degradation over time, with training on classes 0 and 1.}
    \label{fig:model-degradation-over-time-D1}
\end{figure}

The experimental goal is to show the capability of the proposed methodology to correctly assess the model degradation over time when concept drifts are introduced.
For both experiments, the training set is composed by a stratified sample over classes 0 and 1 of about 60\% of records in each class. 
The remaining part of the dataset (40\% class 0, 40\% class 1, the unknown class 2 and/or 3 according to the dataset) is used as test set to assess model degradation.
The first four test sets, from $t_1$ to $t_4$, contain only data known to belong to the classes included in the training set, whereas the last test sets, from $t_5$ to $t_n$, contain both known-class samples and unknown-class samples, in different proportions. 
The test sets have been designed to simulate the flow of time, so subsequent tests extend the data of previous ones, e.g., $t_1$ contains half of the test set for class 0, $t_4$ contains the whole test set for classes 0 and 1, $t_5$ contains all the test samples for classes 0 and 1 and 20\% of the unknown class (class 2 or 3), then at each step another 20\% of the unknown class is added for the evaluation of degradation until $t_n$.

\begin{figure}
    \centering
    \subfloat[Degradation with class 2.]{
    \includegraphics[width=0.45\textwidth]{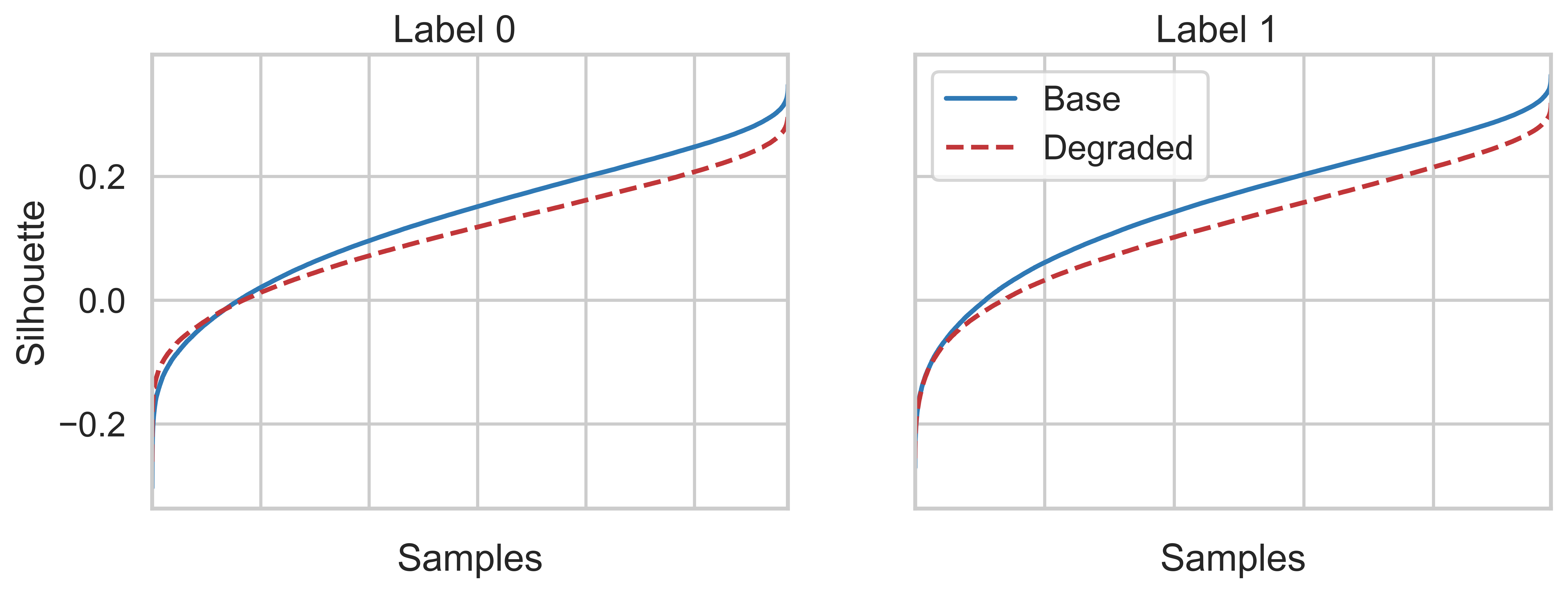}
    \label{fig:d1-sil-test-9}
    }\\
    \subfloat[Degradation with class 3.]{
    \includegraphics[width=0.45\textwidth]{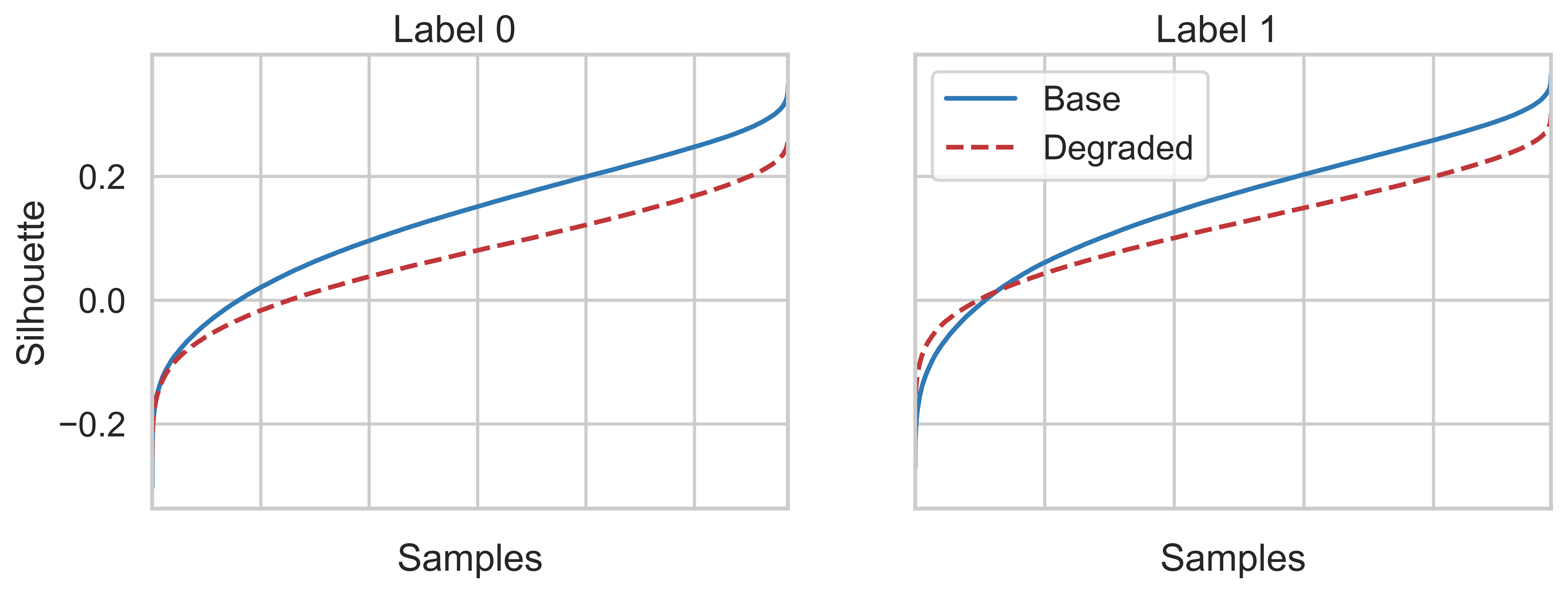}
    \label{fig:d1-sil-test-14}
    }
    \caption{Dataset D1. Baseline and degraded Silhouette curves for each class label at time $t_9$.}
    \label{}
\end{figure}

Figure \ref{fig:model-degradation-over-time-D1} shows the MAAPE degradation percentage for the training classes 0 and 1 in D1 at different time periods, when data belonging to the previously unseen classes 2 (Figure \ref{fig:model-degradation-over-time-1}) and 3 (Figure \ref{fig:model-degradation-over-time-2}) arrive starting from time $t_5$.
While from \textit{$t_1$} to \textit{$t_4$} the overall MAAPE degradation is always below 5\%, from \textit{$t_5$} the overall MAAPE degradation, i.e. the sum of both classes (0 and 1), is constantly higher than 10\%, and the trend is coherent with the introduced drift proportion.
Figures \ref{fig:d1-sil-test-9} and \ref{fig:d1-sil-test-14} report the detailed Silhouette degradation of the known classes 0 and 1 at time \textit{$t_9$}, when there are new samples of the unknown classes 2 and 3 in the test set.

\begin{figure}[!h!]
    \centering
    \includegraphics[width=0.34\textwidth]{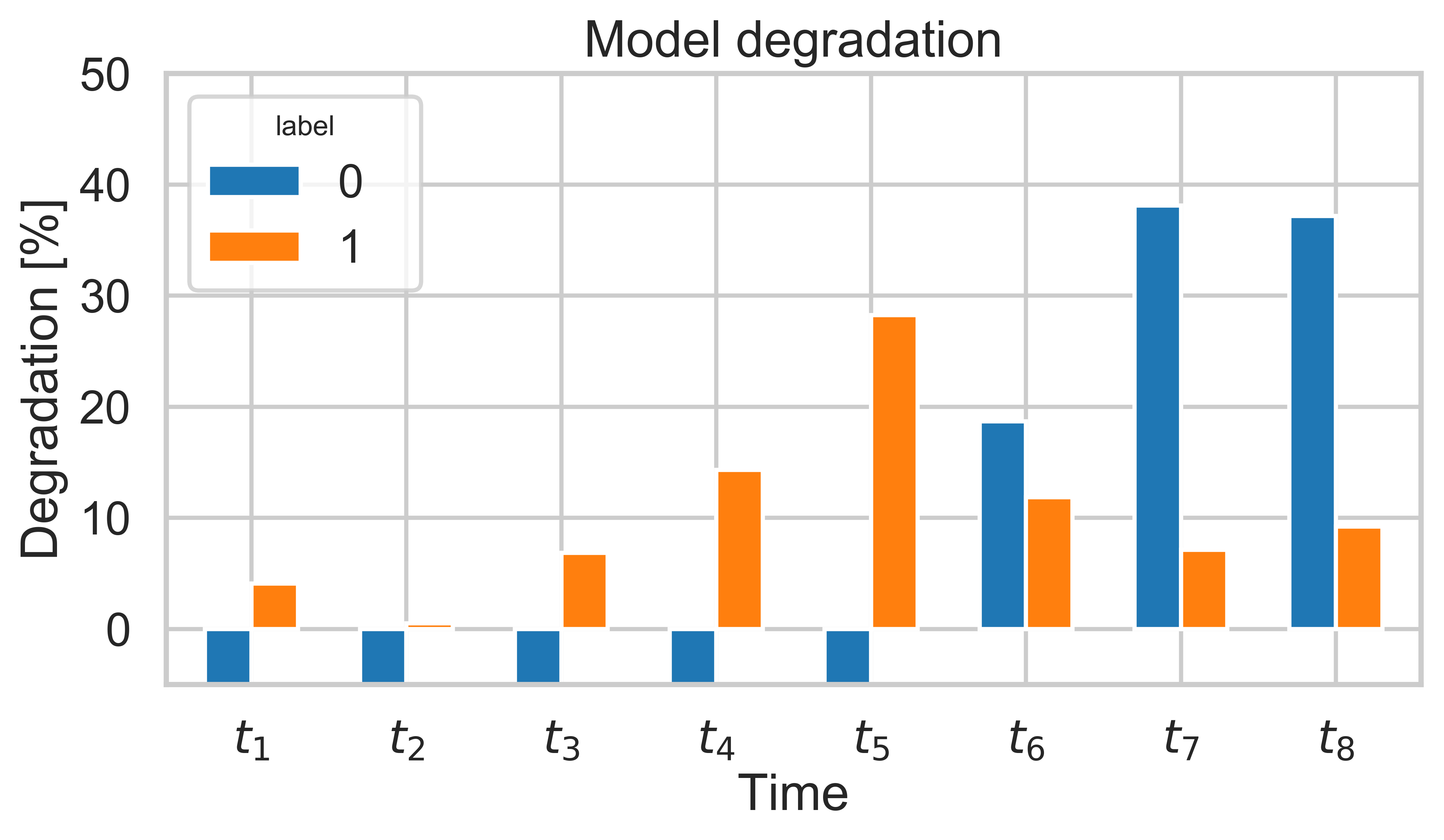}
    \caption{Dataset D2. Model degradation over time, with training on classes 0 (\textit{food-drink}) and 1 (\textit{literature}) and degradation introduced by the new class 2 (\textit{mathematics}).}
    \label{fig:model-degradation-over-time-wiki} 
\end{figure}

Figure \ref{fig:model-degradation-over-time-wiki} shows the average percentage degradation of the Wikipedia document classification model. 
From the histogram, it is possible to see that, up to time \textit{$t_4$}, the new unseen data classified by the model fits well the learnt distribution, with an overall MAAPE below 15\%.
At time \textit{$t_5$}, samples of an unknown class start arriving and the overall MAAPE raises above 28\%.

In both datasets, the sum of the degradation for all known classes correctly detected when the shift was introduced, even if the proportion of new shifted data was low (i.e., 20\% of the unknown class).
Specifically, we note that the overall degradation at least duplicates from the pre-drift ($t_1$ to $t_4$) to the post-drift ($t_5$ to $t_n$), hence proving to be a promising detector of these events.

\section{Conclusions and future works}
\label{sec:conclusion}

This paper presented a novel strategy to detect concept drift, due to the arrival of new unseen samples not fitting the data distribution available at model-training time. 
The proposed approach is based on the evaluation of the predictive model degradation through an unsupervised metric (i.e. the Silhouette index) in a self-evaluating fashion, and shows promising experimental results on two datasets.

Future directions of this research include: 
(i) a comparison with state-of-the-art techniques focusing on drift recognition efficiency,    
(ii) the introduction of alternative unsupervised metrics besides the Silhouette index, 
(iii) the improvement of the self-evaluation triggering mechanism (currently set as a percentage of new data),
and 
(iv) further experiments to assess the generality and the performance of the approach over different real-world datasets presenting known concept drifts.

\bibliographystyle{ACM-Reference-Format}
\bibliography{automl_main.bib}


\begin{thebibliography}{20}


\ifx \showCODEN    \undefined \def \showCODEN     #1{\unskip}     \fi
\ifx \showDOI      \undefined \def \showDOI       #1{#1}\fi
\ifx \showISBNx    \undefined \def \showISBNx     #1{\unskip}     \fi
\ifx \showISBNxiii \undefined \def \showISBNxiii  #1{\unskip}     \fi
\ifx \showISSN     \undefined \def \showISSN      #1{\unskip}     \fi
\ifx \showLCCN     \undefined \def \showLCCN      #1{\unskip}     \fi
\ifx \shownote     \undefined \def \shownote      #1{#1}          \fi
\ifx \showarticletitle \undefined \def \showarticletitle #1{#1}   \fi
\ifx \showURL      \undefined \def \showURL       {\relax}        \fi
\providecommand\bibfield[2]{#2}
\providecommand\bibinfo[2]{#2}
\providecommand\natexlab[1]{#1}
\providecommand\showeprint[2][]{arXiv:#2}

\bibitem[\protect\citeauthoryear{Apiletti, Baralis, Cerquitelli, Garza,
  Giordano, Mellia, and Venturini}{Apiletti et~al\mbox{.}}{2016}]%
        {apiletti2016selina}
\bibfield{author}{\bibinfo{person}{Daniele Apiletti}, \bibinfo{person}{Elena
  Baralis}, \bibinfo{person}{Tania Cerquitelli}, \bibinfo{person}{Paolo Garza},
  \bibinfo{person}{Danilo Giordano}, \bibinfo{person}{Marco Mellia}, {and}
  \bibinfo{person}{Luca Venturini}.} \bibinfo{year}{2016}\natexlab{}.
\newblock \showarticletitle{Selina: a self-learning insightful network
  analyzer}.
\newblock \bibinfo{journal}{\emph{IEEE Transactions on Network and Service
  Management}} \bibinfo{volume}{13}, \bibinfo{number}{3}
  (\bibinfo{year}{2016}), \bibinfo{pages}{696--710}.
\newblock


\bibitem[\protect\citeauthoryear{Apiletti, Barberis, Cerquitelli, Macii, Macii,
  Poncino, and Ventura}{Apiletti et~al\mbox{.}}{2018}]%
        {DBLP:conf/ispa/ApilettiBCMMPV18}
\bibfield{author}{\bibinfo{person}{Daniele Apiletti}, \bibinfo{person}{Claudia
  Barberis}, \bibinfo{person}{Tania Cerquitelli}, \bibinfo{person}{Alberto
  Macii}, \bibinfo{person}{Enrico Macii}, \bibinfo{person}{Massimo Poncino},
  {and} \bibinfo{person}{Francesco Ventura}.} \bibinfo{year}{2018}\natexlab{}.
\newblock \showarticletitle{iSTEP, an Integrated Self-Tuning Engine for
  Predictive Maintenance in Industry 4.0}. In \bibinfo{booktitle}{\emph{{IEEE}
  International Conference on Parallel {\&} Distributed Processing with
  Applications, Ubiquitous Computing {\&} Communications, Big Data {\&} Cloud
  Computing, Social Computing {\&} Networking, Sustainable Computing {\&}
  Communications, ISPA/IUCC/BDCloud/SocialCom/SustainCom 2018, Melbourne,
  Australia, December 11-13, 2018}}. \bibinfo{pages}{924--931}.
\newblock
\urldef\tempurl%
\url{https://doi.org/10.1109/BDCloud.2018.00136}
\showDOI{\tempurl}


\bibitem[\protect\citeauthoryear{Bifet and Gavalda}{Bifet and Gavalda}{2007}]%
        {bifet2007learning}
\bibfield{author}{\bibinfo{person}{Albert Bifet} {and} \bibinfo{person}{Ricard
  Gavalda}.} \bibinfo{year}{2007}\natexlab{}.
\newblock \showarticletitle{Learning from time-changing data with adaptive
  windowing}. In \bibinfo{booktitle}{\emph{Proceedings of the 2007 SIAM
  international conference on data mining}}. SIAM, \bibinfo{pages}{443--448}.
\newblock


\bibitem[\protect\citeauthoryear{Corso, Cerquitelli, and Ventura}{Corso
  et~al\mbox{.}}{2017}]%
        {DBLP:conf/sac/CorsoCV17}
\bibfield{author}{\bibinfo{person}{Evelina~Di Corso}, \bibinfo{person}{Tania
  Cerquitelli}, {and} \bibinfo{person}{Francesco Ventura}.}
  \bibinfo{year}{2017}\natexlab{}.
\newblock \showarticletitle{Self-tuning techniques for large scale cluster
  analysis on textual data collections}. In
  \bibinfo{booktitle}{\emph{Proceedings of the Symposium on Applied Computing,
  {SAC} 2017, Marrakech, Morocco, April 3-7, 2017}}. \bibinfo{pages}{771--776}.
\newblock
\urldef\tempurl%
\url{https://doi.org/10.1145/3019612.3019661}
\showDOI{\tempurl}


\bibitem[\protect\citeauthoryear{Gama, {\v{Z}}liobait{\.e}, Bifet, Pechenizkiy,
  and Bouchachia}{Gama et~al\mbox{.}}{2014}]%
        {gama2014survey}
\bibfield{author}{\bibinfo{person}{Jo{\~a}o Gama}, \bibinfo{person}{Indr{\.e}
  {\v{Z}}liobait{\.e}}, \bibinfo{person}{Albert Bifet}, \bibinfo{person}{Mykola
  Pechenizkiy}, {and} \bibinfo{person}{Abdelhamid Bouchachia}.}
  \bibinfo{year}{2014}\natexlab{}.
\newblock \showarticletitle{A survey on concept drift adaptation}.
\newblock \bibinfo{journal}{\emph{ACM computing surveys (CSUR)}}
  \bibinfo{volume}{46}, \bibinfo{number}{4} (\bibinfo{year}{2014}),
  \bibinfo{pages}{44}.
\newblock


\bibitem[\protect\citeauthoryear{Kim and Kim}{Kim and Kim}{2016}]%
        {MAAPE-kim2016new}
\bibfield{author}{\bibinfo{person}{Sungil Kim} {and} \bibinfo{person}{Heeyoung
  Kim}.} \bibinfo{year}{2016}\natexlab{}.
\newblock \showarticletitle{A new metric of absolute percentage error for
  intermittent demand forecasts}.
\newblock \bibinfo{journal}{\emph{International Journal of Forecasting}}
  \bibinfo{volume}{32}, \bibinfo{number}{3} (\bibinfo{year}{2016}),
  \bibinfo{pages}{669--679}.
\newblock


\bibitem[\protect\citeauthoryear{Klinkenberg and Joachims}{Klinkenberg and
  Joachims}{2000}]%
        {klinkenberg2000detecting}
\bibfield{author}{\bibinfo{person}{Ralf Klinkenberg} {and}
  \bibinfo{person}{Thorsten Joachims}.} \bibinfo{year}{2000}\natexlab{}.
\newblock \showarticletitle{Detecting Concept Drift with Support Vector
  Machines.}. In \bibinfo{booktitle}{\emph{ICML}}. \bibinfo{pages}{487--494}.
\newblock


\bibitem[\protect\citeauthoryear{Lau and Baldwin}{Lau and Baldwin}{2016}]%
        {lau2016empirical}
\bibfield{author}{\bibinfo{person}{Jey~Han Lau} {and} \bibinfo{person}{Timothy
  Baldwin}.} \bibinfo{year}{2016}\natexlab{}.
\newblock \showarticletitle{An empirical evaluation of doc2vec with practical
  insights into document embedding generation}.
\newblock \bibinfo{journal}{\emph{arXiv preprint arXiv:1607.05368}}
  (\bibinfo{year}{2016}).
\newblock


\bibitem[\protect\citeauthoryear{Le and Mikolov}{Le and Mikolov}{2014}]%
        {le2014distributed}
\bibfield{author}{\bibinfo{person}{Quoc Le} {and} \bibinfo{person}{Tomas
  Mikolov}.} \bibinfo{year}{2014}\natexlab{}.
\newblock \showarticletitle{Distributed representations of sentences and
  documents}. In \bibinfo{booktitle}{\emph{International conference on machine
  learning}}. \bibinfo{pages}{1188--1196}.
\newblock


\bibitem[\protect\citeauthoryear{Moreno-Torres, Raeder, Alaiz-Rodr{\'\i}Guez,
  Chawla, and Herrera}{Moreno-Torres et~al\mbox{.}}{2012}]%
        {moreno2012unifying}
\bibfield{author}{\bibinfo{person}{Jose~G Moreno-Torres}, \bibinfo{person}{Troy
  Raeder}, \bibinfo{person}{Roc{\'\i}O Alaiz-Rodr{\'\i}Guez},
  \bibinfo{person}{Nitesh~V Chawla}, {and} \bibinfo{person}{Francisco
  Herrera}.} \bibinfo{year}{2012}\natexlab{}.
\newblock \showarticletitle{A unifying view on dataset shift in
  classification}.
\newblock \bibinfo{journal}{\emph{Pattern Recognition}} \bibinfo{volume}{45},
  \bibinfo{number}{1} (\bibinfo{year}{2012}), \bibinfo{pages}{521--530}.
\newblock


\bibitem[\protect\citeauthoryear{Pedregosa, Varoquaux, Gramfort, Michel,
  Thirion, Grisel, Blondel, Prettenhofer, Weiss, Dubourg, Vanderplas, Passos,
  Cournapeau, Brucher, Perrot, and Duchesnay}{Pedregosa et~al\mbox{.}}{2011}]%
        {scikit-learn}
\bibfield{author}{\bibinfo{person}{F. Pedregosa}, \bibinfo{person}{G.
  Varoquaux}, \bibinfo{person}{A. Gramfort}, \bibinfo{person}{V. Michel},
  \bibinfo{person}{B. Thirion}, \bibinfo{person}{O. Grisel},
  \bibinfo{person}{M. Blondel}, \bibinfo{person}{P. Prettenhofer},
  \bibinfo{person}{R. Weiss}, \bibinfo{person}{V. Dubourg}, \bibinfo{person}{J.
  Vanderplas}, \bibinfo{person}{A. Passos}, \bibinfo{person}{D. Cournapeau},
  \bibinfo{person}{M. Brucher}, \bibinfo{person}{M. Perrot}, {and}
  \bibinfo{person}{E. Duchesnay}.} \bibinfo{year}{2011}\natexlab{}.
\newblock \showarticletitle{Scikit-learn: Machine Learning in {P}ython}.
\newblock \bibinfo{journal}{\emph{Journal of Machine Learning Research}}
  \bibinfo{volume}{12} (\bibinfo{year}{2011}), \bibinfo{pages}{2825--2830}.
\newblock


\bibitem[\protect\citeauthoryear{Proto, Corso, Ventura, and Cerquitelli}{Proto
  et~al\mbox{.}}{2018}]%
        {DBLP:conf/bigdata/ProtoCVC18}
\bibfield{author}{\bibinfo{person}{Stefano Proto}, \bibinfo{person}{Evelina~Di
  Corso}, \bibinfo{person}{Francesco Ventura}, {and} \bibinfo{person}{Tania
  Cerquitelli}.} \bibinfo{year}{2018}\natexlab{}.
\newblock \showarticletitle{Useful ToPIC: Self-Tuning Strategies to Enhance
  Latent Dirichlet Allocation}. In \bibinfo{booktitle}{\emph{2018 {IEEE}
  International Congress on Big Data, BigData Congress 2018, San Francisco, CA,
  USA, July 2-7, 2018}}. \bibinfo{pages}{33--40}.
\newblock
\urldef\tempurl%
\url{https://doi.org/10.1109/BigDataCongress.2018.00012}
\showDOI{\tempurl}


\bibitem[\protect\citeauthoryear{Rousseeuw}{Rousseeuw}{1987}]%
        {Rousseeuw1987}
\bibfield{author}{\bibinfo{person}{Peter~J. Rousseeuw}.}
  \bibinfo{year}{1987}\natexlab{}.
\newblock \showarticletitle{Silhouettes: A graphical aid to the interpretation
  and validation of cluster analysis}.
\newblock \bibinfo{journal}{\emph{J. Comput. Appl. Math.}}
  \bibinfo{volume}{20} (\bibinfo{year}{1987}), \bibinfo{pages}{53 -- 65}.
\newblock
\showISSN{0377-0427}


\bibitem[\protect\citeauthoryear{{Roveri}}{{Roveri}}{2019}]%
        {8618443}
\bibfield{author}{\bibinfo{person}{M. {Roveri}}.}
  \bibinfo{year}{2019}\natexlab{}.
\newblock \showarticletitle{Learning Discrete-Time Markov Chains Under Concept
  Drift}.
\newblock \bibinfo{journal}{\emph{IEEE Transactions on Neural Networks and
  Learning Systems}} (\bibinfo{year}{2019}), \bibinfo{pages}{1--13}.
\newblock
\showISSN{2162-237X}
\urldef\tempurl%
\url{https://doi.org/10.1109/TNNLS.2018.2886956}
\showDOI{\tempurl}


\bibitem[\protect\citeauthoryear{{Sun}, {Tang}, {Zhu}, and {Yao}}{{Sun}
  et~al\mbox{.}}{2018}]%
        {8246541}
\bibfield{author}{\bibinfo{person}{Y. {Sun}}, \bibinfo{person}{K. {Tang}},
  \bibinfo{person}{Z. {Zhu}}, {and} \bibinfo{person}{X. {Yao}}.}
  \bibinfo{year}{2018}\natexlab{}.
\newblock \showarticletitle{Concept Drift Adaptation by Exploiting Historical
  Knowledge}.
\newblock \bibinfo{journal}{\emph{IEEE Transactions on Neural Networks and
  Learning Systems}} \bibinfo{volume}{29}, \bibinfo{number}{10}
  (\bibinfo{date}{Oct} \bibinfo{year}{2018}), \bibinfo{pages}{4822--4832}.
\newblock
\showISSN{2162-237X}
\urldef\tempurl%
\url{https://doi.org/10.1109/TNNLS.2017.2775225}
\showDOI{\tempurl}


\bibitem[\protect\citeauthoryear{Tan, Steinbach, and Kumar}{Tan
  et~al\mbox{.}}{2005}]%
        {Tan:2005:IDM:1095618}
\bibfield{author}{\bibinfo{person}{Pang-Ning Tan}, \bibinfo{person}{Michael
  Steinbach}, {and} \bibinfo{person}{Vipin Kumar}.}
  \bibinfo{year}{2005}\natexlab{}.
\newblock \bibinfo{booktitle}{\emph{Introduction to Data Mining, (First
  Edition)}}.
\newblock \bibinfo{publisher}{Addison-Wesley Longman Publishing Co., Inc.},
  \bibinfo{address}{Boston, MA, USA}.
\newblock
\showISBNx{0321321367}


\bibitem[\protect\citeauthoryear{Tsymbal}{Tsymbal}{2004}]%
        {tsymbal2004problem}
\bibfield{author}{\bibinfo{person}{Alexey Tsymbal}.}
  \bibinfo{year}{2004}\natexlab{}.
\newblock \showarticletitle{The problem of concept drift: definitions and
  related work}.
\newblock \bibinfo{journal}{\emph{Computer Science Department, Trinity College
  Dublin}} \bibinfo{volume}{106}, \bibinfo{number}{2} (\bibinfo{year}{2004}),
  \bibinfo{pages}{58}.
\newblock


\bibitem[\protect\citeauthoryear{Ventura, Proto, Apiletti, Cerquitelli,
  Panicucci, Baralis, Macii, and Macii}{Ventura et~al\mbox{.}}{2019}]%
        {ventura2019scale}
\bibfield{author}{\bibinfo{person}{Francesco Ventura}, \bibinfo{person}{Stefano
  Proto}, \bibinfo{person}{Daniele Apiletti}, \bibinfo{person}{Tania
  Cerquitelli}, \bibinfo{person}{Simone Panicucci}, \bibinfo{person}{Elena
  Baralis}, \bibinfo{person}{Enrico Macii}, {and} \bibinfo{person}{Alberto
  Macii}.} \bibinfo{year}{2019}\natexlab{}.
\newblock \showarticletitle{A new unsupervised predictive-model self-assessment
  approach that SCALEs}. In \bibinfo{booktitle}{\emph{2019 IEEE International
  Congress on Big Data (BigData Congress)}}. IEEE, \bibinfo{pages}{144--148}.
\newblock
\urldef\tempurl%
\url{https://doi.org/10.1109/BigDataCongress.2019.00033}
\showDOI{\tempurl}


\bibitem[\protect\citeauthoryear{{Vorburger} and {Bernstein}}{{Vorburger} and
  {Bernstein}}{2006}]%
        {4053163}
\bibfield{author}{\bibinfo{person}{P. {Vorburger}} {and} \bibinfo{person}{A.
  {Bernstein}}.} \bibinfo{year}{2006}\natexlab{}.
\newblock \showarticletitle{Entropy-based Concept Shift Detection}. In
  \bibinfo{booktitle}{\emph{Sixth International Conference on Data Mining
  (ICDM'06)}}. \bibinfo{pages}{1113--1118}.
\newblock
\showISSN{1550-4786}
\urldef\tempurl%
\url{https://doi.org/10.1109/ICDM.2006.66}
\showDOI{\tempurl}


\bibitem[\protect\citeauthoryear{{Wang}, {Minku}, and {Yao}}{{Wang}
  et~al\mbox{.}}{2018}]%
        {8246564}
\bibfield{author}{\bibinfo{person}{S. {Wang}}, \bibinfo{person}{L.~L. {Minku}},
  {and} \bibinfo{person}{X. {Yao}}.} \bibinfo{year}{2018}\natexlab{}.
\newblock \showarticletitle{A Systematic Study of Online Class Imbalance
  Learning With Concept Drift}.
\newblock \bibinfo{journal}{\emph{IEEE Transactions on Neural Networks and
  Learning Systems}} \bibinfo{volume}{29}, \bibinfo{number}{10}
  (\bibinfo{date}{Oct} \bibinfo{year}{2018}), \bibinfo{pages}{4802--4821}.
\newblock
\showISSN{2162-237X}
\urldef\tempurl%
\url{https://doi.org/10.1109/TNNLS.2017.2771290}
\showDOI{\tempurl}


\end{thebibliography}

\end{document}